# Medical Diagnosis with Large Scale Multimodal Transformers: Leveraging Diverse Data for More Accurate Diagnosis

Authors: Firas Khader (1), Gustav Müller-Franzes (1), Tianci Wang (1), Tianyu Han (2), Soroosh Tayebi Arasteh (1), Christoph Haarburger (3), Johannes Stegmaier (4), Keno Bressem (5), Christiane Kuhl (1), Sven Nebelung (1), Jakob Nikolas Kather* (6, 7, 8, 9), Daniel Truhn* (1)

*equal contribution

(1) Department of Diagnostic and Interventional Radiology, University Hospital Aachen, Germany
(2) Physics of Molecular Imaging Systems, Experimental Molecular Imaging, RWTH Aachen University, Germany
(3) Ocumeda GmbH, Germany
(4) Institute of Imaging and Computer Vision, RWTH Aachen, Germany
(5) Department of Diagnostic and Interventional Radiology, Charité – Universitätsmedizin Berlin, corporate member of Freie Universität Berlin and Humboldt-
(6) Universität zu Berlin, Germany
(7) Department of Medicine III, University Hospital Aachen, Germany
(8) Else Kroener Fresenius Center for Digital Health, Medical Faculty Carl Gustav Carus, Technical University Dresden, Germany
(9) Division of Pathology and Data Analytics, Leeds Institute of Medical Research at St James's, University of Leeds, UK.
(10) Medical Oncology, National Center for Tumor Diseases (NCT), University Hospital Heidelberg, Germany

# Abstract

Multimodal deep learning has been used to predict clinical endpoints and diagnoses from clinical routine data. However, these models suffer from scaling issues: they have to learn pairwise interactions between each piece of information in each data type, thereby escalating model complexity beyond manageable scales. This has so far precluded a widespread use of multimodal deep learning.

Here, we present a new technical approach of "learnable synergies", in which the model only selects relevant interactions between data modalities and keeps an "internal memory" of relevant data. Our approach is easily scalable and naturally adapts to multimodal data inputs from clinical routine. We demonstrate this approach on three large multimodal datasets from radiology and ophthalmology and show that it outperforms state-of-the-art models in clinically relevant diagnosis tasks.

Our new approach is transferable and will allow the application of multimodal deep learning to a broad set of clinically relevant problems.

# Introduction

In medicine, the diagnosis of a disease inherently builds upon data from multiple sources. A clinician will base decisions on radiological images, on clinical data, on the history of the patient, on laboratory findings and on many additional modalities. The human mind is capable of condensing all of these inputs into a rational decision. It has long been proposed that deep learning (DL) can assist medical doctors in certain tasks and this is certainly true given the plethora of research demonstrating the performance of deep learning models that rival - or even surpass - the performance of human experts.[1] However, there is one crucial impediment that limits the general applicability of such models: they are almost exclusively tailored to solve tasks in one kind of data at a time - be it diagnosis of pathologies in radiological images[2,3], detection of genetic alterations in histopathological images[4] or other, mostly image-based tasks. This is a serious shortcoming as it leaves the task of information condensation from multiple sources to the human expert. However, medicine is becoming more complex with ever increasing amounts of data being considered even for the diagnosis of single patients.[5] Models that are capable of combining both image and non-image data as inputs are needed to truly support physicians in decision-making[6]. The prevailing deep learning architectures of the past are not suited to deal with large image and non-image data: convolutional neural networks (CNNs) make use of intrinsic biases that build on image properties, such as correlations between neighboring pixels and integration of non-image information is not straightforward.[7] In the non-medical domain, transformers have recently been shown to be competitive to CNNs in image processing while simultaneously being ideally suited to combine image and non-image data as they treat both modalities on equal footings.[8,9] Thus, their application in medicine is the next step.[10,11] However, transformers are not easily scalable: Medical datasets are large and computational load for transformer models scales quadratically with the amount of inputs. Without remedy this will limit progress in medical research. As an exemplary case, consider a model trained to recognize a life threatening change of the patient's health state in the intensive care unit. Intensive care patients are continuously monitored and such a model might take in the arterial blood pressure measurements as inputs. The most knowledgeable model will account for all of the blood pressure measurements of that patient in the past. But even extending the input reach of the model from the last minute to the last hour would increase the required computational resources by a factor of approximately 3,600. We solve this problem by utilizing a new technique based on the transformer architecture that scales linearly in the input, thus allowing us to train much bigger multimodal models. We demonstrate its capabilities by applying our architecture to a public dataset of intensive care patients and we confirm our findings on two independent international datasets of intensive care patients and of walk-in patients being screened for ophthalmological diseases, see **Figure 1**. In combining both image and non-image inputs, we show that we surpass the state of the art while simultaneously preserving linear scalability. Thus opening the path to the development of truly big and robust multimodal DL models.

# Results

**Multimodal Transformers can Diagnose Multiple Diseases**

In order to address the need for a scalable multimodal transformer we propose a novel architecture that uses both image and non-image data as inputs and that is flexible in the amount of clinical parameters serving as inputs to the model, see Figure **2A**. We test our model on publicly available data of n=36,542 patients undergoing intensive care treatment[12,13]. We employ chest radiographs and accompanying clinical data as inputs to the model and let the model predict a comprehensive set of 25 clinical conditions. We find that the area under the receiver-operator characteristic (AUC) increases consistently if both image and non-image data are employed as compared to the setup that uses either image or non-image data only, see Table **1**. On average, the AUC increased to 0.77 [95% CI: 0.77, 78] when using both chest radiographs and clinical data as compared to 0.70 [95% CI: 0.69, 0.71] when using only chest radiographs and as compared to 0.72 [95% CI: 0.72, 0.73] when using only clinical data. More importantly, the multimodal transformer model surpasses conventional convolutional neural networks, while simultaneously allowing for a flexible number of input parameters, see Table **2** for a detailed comparison to the previous state of the art performance achieved with CNNs. To demonstrate the generalizability of our model, we further evaluate the model on two additional tasks and on independent datasets: 1) Comprehensive radiological diagnosis of chest radiographs based on image data and accompanying laboratory data[14] and 2) Anomaly detection for patients undergoing an ophthalmological screening via fundoscopy with accompanying anamnestic data. For 1) we find that the average AUC increases to 0.84 [95% CI: 0.83, 0.84] when using chest radiographs and clinical data, compared to 0.83 [95% CI: 0.82, 0.83] when using only chest radiographs and to 0.67 [95% CI: 0.66, 0.67] when using only clinical parameters. For 2) we find that the average AUC increases to 0.83 [95% CI: 0.79, 0.86] when using funduscopic and anamnestic data, as compared to 0.80 [95% CI: 0.77, 0.84] when using only fundoscopy and as compared to 0.71 [95% CI: 0.66, 0.75] when using only clinical history, see **Figure 3**. We thus find that multimodal transformers are a viable architecture to increase diagnostic accuracy when fed with multimodal data.

**Multimodal Transformers Scale to Large Real-World Datasets**

In clinical applications, multimodal deep learning models will most likely incorporate as much information about the patient as possible to arrive at the most accurate diagnoses. For conventional transformer architectures this is a problem since the computational complexity scales quadratically in the number of inputs.[15] To keep deep learning models trainable even if they comprise all available patient data, it is necessary to switch to a more favorable scaling behavior and we therefore set up our model to build on cross-attention in combination with self-attention to retain a linear scaling behavior. We tested how much time it took to train a model on medical data with varying amounts of input data. We measured the time needed to train our novel transformer architecture for one epoch and compared it to a conventional transformer architecture, where image tokens and clinical parameter tokens are concatenated and fed through a common transformer encoder in all stages. We varied the amount of input data by setting the number of time points at which the clinical parameters were measured to 100, 200, 400, 800, 1600 and 3200. We found that the time it took to train the conventional model greatly exceeded the time it took to train our scalable model, see Figure **2B**. Similarly important, we found that GPU memory usage escalates for the conventional model, while our model retains a low memory footprint even for a high number of data inputs, see Figure **2C**. Together, these data show that multimodal training of transformer models on big data requires architectures that do not explode in terms of training time and GPU memory requirements. The former is important to train on large datasets (i.e., number of patients is large), while the latter is important to include samples that each contain large data (i.e., number of data per patient is large). With our proposed model, we ensure benevolent behavior in both regimes and thus provide a blueprint for the development of large comprehensive multimodal deep learning models.

**Multimodal Transformers are Inherently Robust to Missing Data**

To account for a typical clinical situation in which input data is partly missing (e.g., a clinical test might be available for some patients while that test might not have been done for other patients), all practical deep learning models that are utilized in clinical routine need to be robust towards missing data. We demonstrate that the proposed transformer architecture works when data is missing and resembles human reasoning in the sense that performance drops continuously when increasing amounts of clinically relevant data are missing. We proceed by feeding patient data from the test set to the trained transformer with parts of the input parameters randomly dropped. As shown in **Figure 4D**, the performance in terms of the average AUC drops continuously when growing amounts of data are deleted in agreement with expectations.

We thus confirm that multimodal transformers behave analogously to humans in that they base their diagnosis on all available data and that the accuracy of that diagnosis decreases continuously when less data is available.

**Multimodal Transformers can Rediscover Pathophysiology**

While deep learning models are trained on big data to achieve human performance, their internal reasoning process often remains elusive. To uncover relationships between available data and diagnostic accuracy we make use of a new method that employs a quantitative measure of how much information is lost when a dedicated type of input data (e.g., blood glucose levels) is left out. We proceed by calculating the mutual information between the model's prediction and the ground truth over all patients from the test set. This calculation is performed twice: once with all inputs given to the model and a second time with the same inputs except for the input type that is to be examined. We find that this analysis largely agrees with clinical reasoning and that clinical parameters that are relevant to a specific patient state - such as blood pressure for shock or glucose concentration for diabetes - lead to the greatest information loss when left out, see **Figure 4E**. A comprehensive overview for all clinical conditions is given in the appendix (**Supplementary Figure 1)**. This paradigm thus allows us to quantitatively uncover the relationship between diagnostic tests and a loss in diagnostic accuracy if these tests are left out. This may help to steer clinical routine towards a more efficient and economic use of diagnostic tests.

**Multimodal Transformer are Explainable**

To gain insight into the inner workings of CNNs, various techniques have been proposed over the years[16,17]. For multimodal models based on CNNs, these visualization techniques can not be extended in a straightforward way due to the mixing with non-image data[16]. For transformer architectures, however, the inherent attention mechanism[18] allows for a straight-forward extension that lends itself to direct interpretability: the attention score denotes which parts of the image are given greater weight during the decision making process and thus its visualization gives direct insights into the inner workings of the transformer model. In Figure **4 A, B and C** we show illustrative attention maps projected onto the input space of the image input using the attention-rollout mechanism[19]. We find that for all datasets and image domains (i.e. chest radiographs and fundoscopy images) the model focuses on relevant parts of the image. We thus confirm that the attention mechanism of transformers can be employed to explain the neural networks' image-based reasoning in the multimodal setup.

# Discussion

In recent years there has been a surge of applications of deep learning models to medical problems.[14,20–22] These models mainly work on one data modality, e.g., image data, only. However, in clinical practice almost every diagnostic decision is based on a wide variety of inputs. This can include information from imaging tests such as X-rays or MRIs, laboratory results from blood or tissue samples, genetic information and the patient's medical history. This information helps doctors to make an accurate diagnosis and to develop a treatment plan that is tailored to the individual patient. All of these data sources contribute to clinical reasoning and hence there has been a growing interest in the application of deep learning models that are capable of processing multimodal information: Boehm et al. demonstrated that multimodal data integration can help to stratify cancer patients[23] and Vanguri et al. showed that the integration of radiological, pathological and genomic data in deep learning models can predict therapy response for patients with non-small cell lung cancer[24]. Because these studies have been performed on a comparatively small number of patients (n<500), they relied on manual segmentations and the extraction of handcrafted features with a subsequent combination in shallow networks. Training of multimodal models on big data requires the introduction of new architectures. Transformer models have been proposed as an ideal candidate for such multimodal reasoning, as they have first been developed on non-image data[25,26] and have now proven competitive to CNNs on image data.[27] In the present study, we developed the first scalable, fully transformer-based approach for multimodal prediction on medical image and non-image data. Our model is more accurate than existing approaches, is robust to missing data and allows insights into the network's decision making process. Most importantly though, our model is scalable to be applied on datasets in which both the number of patients and the data per patient is large. Both are equally important since such models will most likely be most valuable in application scenarios where the available data is complex or difficult to interpret (i.e. a lot of data per patient) and where a large number of patients is needed to train such complex models.

We have tested the capabilities of the proposed transformer model on three distinct datasets: first intensive care patients from the US with a wide variety of clinical tests performed on them. Second, an independent dataset of intensive care patients from Germany with associated laboratory values only and third, a population of patients undergoing screening for ophthalmological diseases with accompanying anamnestic data. In all three datasets we found a consistent increase in diagnostic accuracy when utilizing accompanying non-image clinical data next to the image data. We additionally demonstrated how the trained models can be employed to extract information about the reasoning of these networks both on the image and the non-image domain and how this might help to uncover clinically meaningful information.

Our model is constructed such that it can also be applied on truly large multimodal datasets. However, the availability of such public or private multimodal datasets is still limited and we have not found any accessible dataset with a size of greater than one million patients. Future work should therefore demonstrate the applicability of our architecture on such data. Our work has further limitations that give rise to future research: first, the cases demonstrated here make use of two-dimensional imaging data. A substantial amount of medical image data is three-dimensional though and it should be tested if the presented paradigms hold on three-dimensional data once such datasets become available. Second, we tested our network in the context of supervised learning. This requires the presence of labels for each patient and limits the range of data that can be employed to train our architecture. Third, domain-transferability of the model could not be tested due to the lack of suitable datasets that have concordant labels and concordant data available for training.

We hope that our work triggers further research into these areas and that it promotes the clinical applicability of multimodal deep learning models.

# Online Methods

**Ethics Statement**

All experiments were conducted in accordance with the Declaration of Helsinki and the International Ethical Guidelines for Biomedical Research Involving Human Subjects by the Council for International Organizations of Medical Sciences (CIOMS). The study has additionally been approved by the local ethical committee (EK 22-319).

**Description of the Datasets**

To allow for replication of our results and to foster research in this direction, we evaluate our model primarily on the publicly available MIMIC (Medical Information Mart for Intensive Care) database[12,13]. This database comprises retrospectively collected image and non-image data of over 40,000 patients admitted to an intensive care unit or the emergency department at the Beth Israel Deaconess Medical Center between 2008 and 2019. We follow the work of Hayat et al.[28] and extract imaging and non-imaging information from the MIMIC-IV[12] and MIMIC-CXR-JPG[13] database resulting in a subset of 45,676 samples from n=36,542 patients (mean age, 63 years ± 17 [SD], 56% male) that either contained information about 15 clinical parameters (i.e., systolic blood pressure, respiratory rate, motor response in Glasgow Coma Scale, oxygen inspiration, verbal response in Glasgow Coma Scale, heart rate, response to the command of opening the eyes in Glasgow Coma Scale, body temperature, body weight, mean blood pressure, acidic value of blood serum, diastolic blood pressure, blood glucose, body height, blood oxygen saturation) only or in addition also contained imaging information in form of chest radiographs. Additionally, we evaluate our model on an in-house dataset of n=45,016 patients (mean age, 66 years ± 16 [SD], 61% male) that were admitted to an intensive care unit of a tertiary academic medical center (University Hospital Aachen) between 2009 and 2020.[14] In addition to imaging data, i.e. chest radiographs, this dataset also contains time-series data of laboratory values, i.e., C-reactive protein (CRP), leukocytes, (LEUK), Procalcitonin (PCT) and Brain Natriuretic Peptide (BNP). This dataset was manually annotated by 98 trained radiologists and contains binary labels for pleural effusion (left and right), atelectasis (left and right), pulmonary opacities (left and right), pulmonary congestion and cardiomegaly. To show the capabilities of our model in also handling data from a different domain, we use an additional internal dataset of n=1,930 patients (mean age, 56 years ± 16 [SD]) that consists of fundoscopy images and accompanying non-imaging parameters (i.e., age, history of high blood pressure, glaucoma presence within family as well as eyesight and pressure on both eyes) of ophthalmology walk-ins acquired between January 2022 and October 2022. Two ophthalmologists then labeled the images as either "require imminent or immediate control by ophthalmologist" or "no control is required".

**Data Preprocessing**

We proceed by splitting the MIMIC dataset into a training set of 42,628 samples (n=33,893 patients), a validation set of 882 samples (n=740 patients) and test set of 2,166 samples (n=1,909 patients) as is done by Hayat et al.[28], where the training and validation set include partial data, meaning that imaging information may be missing, whereas the test set contains imaging and non-imaging information for all samples. All images were normalized to the range between 0 and 255 and a histogram equalization was performed to enhance the image contrast. This procedure has been performed by the dataset providers. We then proceed by resizing the images to an equal width and height of 384 pixels to match the resolution used for the pretrained ImageNet backbones. We perform a similar preprocessing routine, i.e., normalization, histogram equalization and resizing on the internal dataset of chest radiographs and fundoscopy images and split the datasets into a training, validation and test set. The internal dataset of chest radiographs consisting of 193,556 samples (n=45,016 patients) is thus split into a training set of 122,294 samples (n=28,809 patients), validation set of 31,243 samples (n=7,203 patients) and a test set of 40,028 samples (n=9,004 patients). Similarly, the the fundoscopy dataset comprised of 3,860 samples (n=1,930 patients) is split into training set of 2,586 samples (n=1,293 patients), a validation set of 502 samples (n=251 patients) and a test set of 772 samples (n=386 patients). As a final step, a z-normalization was performed on all images to match the dataset statistics (mean and standard deviation) of the ImageNet[29] dataset to leverage the pre-trained backbones in the model.

**Architecture**

Our architecture is based on the transformer architecture[18] that uses the attention mechanism to extract features from the input space. In broad terms, we employ a vision transformer[9] (ViT) backbone for image processing and utilize the cross-attention mechanism for the processing of non-image information[18,30]. This serves to keep the non-image input scalable (the number of non-image inputs might be very large, e.g., a patient might have 20 blood samples taken during the hospital stay) and flexible (the number of non-image inputs might vary, e.g., patient A might have 20 blood samples taken, while patient B had only 5). While the traditional transformer architecture scales with $N^2$ where N is a fixed number of input tokens (i.e., the number of image patches or non-imaging parameters used), our architecture scales linearly in N with N being a flexible number of input tokens as we leverage the cross-attention mechanism[30] such that only a pre-specified set of latent tokens attend to the N input tokens in comparison to having all input tokens attend to all other input tokens.

Thus, the conventional transformer model is only capable of integrating information of clinical parameters for a fixed, small number of time steps, while we design our model architecture to support inputs with a variable amount of time steps. More precisely, let $x_{CXR} \in \mathbb{R}^{H \times W \times C}$ denote an image, in this case a chest radiograph, where $H$ denotes the image height, $W$ denotes the image width and $C$ denotes the number of channels. We first divide this image into non-overlapping patches of size 32 x 32 and subsequently feed the image through a learnable linear projection layer to create image tokens $z_{CXR} \in \mathbb{R}^{N \times D_{CXR}}$, where N denotes the number of image tokens and $D_{CXR}$ denotes the dimension of the latent representation ($D_{CXR} = 384$ in our case). A positional encoding of the same dimensionality is added to the latent representation. A transformer encoder block is then used to process the imaging information. Here, we used a vision transformer that supports images of size $384 \times 384$, which had previously been pre-trained on ImageNet[29]. To process the clinical parameters $x_{CP}$, we utilize cross-attention as proposed in the perceiver model[30]. In detail, cross-attention modules are used to inject information from a large number of input tokens carrying the information about the clinical parameters at different time points. This information is condensed in a set of $P$ tokens of dimension $D_{CP}$ that are carried through the transformer architecture as depicted in **Figure 2A**. The initialization of these tokens is learnable. In each block, information about the clinical parameters is injected by first adding a fixed positional encoding (in this case encoding time) to the set of clinical parameters and then computing a set of value and key vectors to which the set of $P$ tokens can attend to. Because we can now freely set the number of learnable tokens, we opt for a number that is much smaller than the largest sequence length and thus reduce the computational burden in the transformer blocks. A linear projection layer is then used to map the dimensionality of the latent tokens $D_{CP}$ to that of the image tokens $D_{CXR}$. The resulting latent tokens are then concatenated with the output of the transformer encoder used to process the images and a fixed positional encoding is added to each token to allow for a differentiation between imaging and non-imaging tokens. Additionally, a class token of the same dimensionality is prepended to the initial set of tokens. A final transformer encoder is then used to process the resulting tokens and therefore allows for cross-modality information fusion. A final multi-layer perceptron in combination with a Sigmoid activation function[31] is then used to form the final prediction scores (i.e., the probability for a certain disease).

To demonstrate the capabilities of our model in handling non-imaging parameters $x_{CP}$ of a different format, we train our models using a different format for each dataset: (1) In the case of the MIMIC dataset, we follow the data format suggested by Hayat et al.[28], where the input $x_{CP} \in \mathbb{R}^{K \times T}$ denotes a matrix with $K$ clinical parameter items and $T$ number of timesteps. Each parameter vector for a specified time-step $t$ contains the information about the clinical parameters and, when one of the clinical parameters is missing, the imputed value according to **Supplementary Table 1**. Additionally, information about whether a certain value has been imputed is appended to this vector in the form of binary flags for each clinical parameter (i.e., 0 if it has been imputed and 1 if not). (2) The data-format as used for the MIMIC dataset has one notable shortcoming: an impute value is needed for every single clinical parameter. Therefore, we propose a different data-format for the internal dataset of chest radiographs and clinical parameters, where instead we provide information about the clinical parameters in the form $x_{CP} \in \mathbb{R}^{J \times G}$. $G$ denotes the number of available clinical parameters over all timesteps and $J$ denotes the vector length needed to encode the information about the clinical parameter. In particular, given the value of a certain type of parameter (i.e., CRP, LEUK, BNP or PCT) and the temporal difference to the acquisition time of the chest radiograph, the vector of size $J$ is formed by concatenating the numerical value, the type of clinical parameter as represented by a one-hot encoding and the temporal difference in terms of Fourier features[18]. (3) To show the applicability of our model to non-time-series data, we propose another encoding as is used for the dataset of fundoscopy images and the corresponding non-imaging parameters. In this case, we suggest to formulate $x_{CP}$ as an element of $\mathbb{R}^{1 \times F}$, where $F$ denotes the number of non-imaging parameters available, resulting in a simple row vector, where each entry indicates the numerical value of the corresponding non-imaging parameter.

In order to present a diverse set of inputs to the model, we augment the training data by flipping the images horizontally, randomly rotating the image to ± 45° degrees, scaling the image by ± 15% and performing random horizontal and vertical translations (± 15%). Additionally, numerical non-imaging parameters are augmented by sampling a new value from a Gaussian with standard deviation 0.1, centered at the numeric value. All models were trained on an NVIDIA Quadro RTX 6000.

**Data availability**

The MIMIC dataset, including imaging and non-imaging data, is publicly available via PhysioNet[32]. The other two datasets are private due to data protection issues, but will be shared by the authors upon submission of a research proposal and given the consent of the data protection officer and the ethical board.

**Code availability**

The code used to train our model is publicly available on GitHub: https://github.com/FirasGit/lsmt.

# Author contributions

The experiment was designed by FK, JNK, SN and DT. The model architecture was set up by FK, JNK and DT. The code was written by FK. Statistical analyses were performed by FK, SN, and DT. All authors contributed to the interpretation of the results, the writing of the final manuscript, and agreed to submission of this paper.

# Acknowledgements

We thank the MIMIC consortium and the Ocumeda AG for providing the datasets used in our study.

# Funding

JNK is supported by the German Federal Ministry of Health (DEEP LIVER, ZMVI1-2520DAT111) and the Max-Eder-Programme of the German Cancer Aid (grant #70113864). SN is supported by the Deutsche Forschungsgemeinschaft (DFG) (No. NE 2136/3-1).

# Disclosures

For transparency, we provide the following information: JNK declares consulting services for Owkin, France, Panakeia, UK, and DoMore Diagnostics, Norway. KB declares speaker fees for Canon Medical Systems Cooperation, Japan. CH is a shareholder of Ocumeda AG, Switzerland.

# Figures

**Study Cohort**

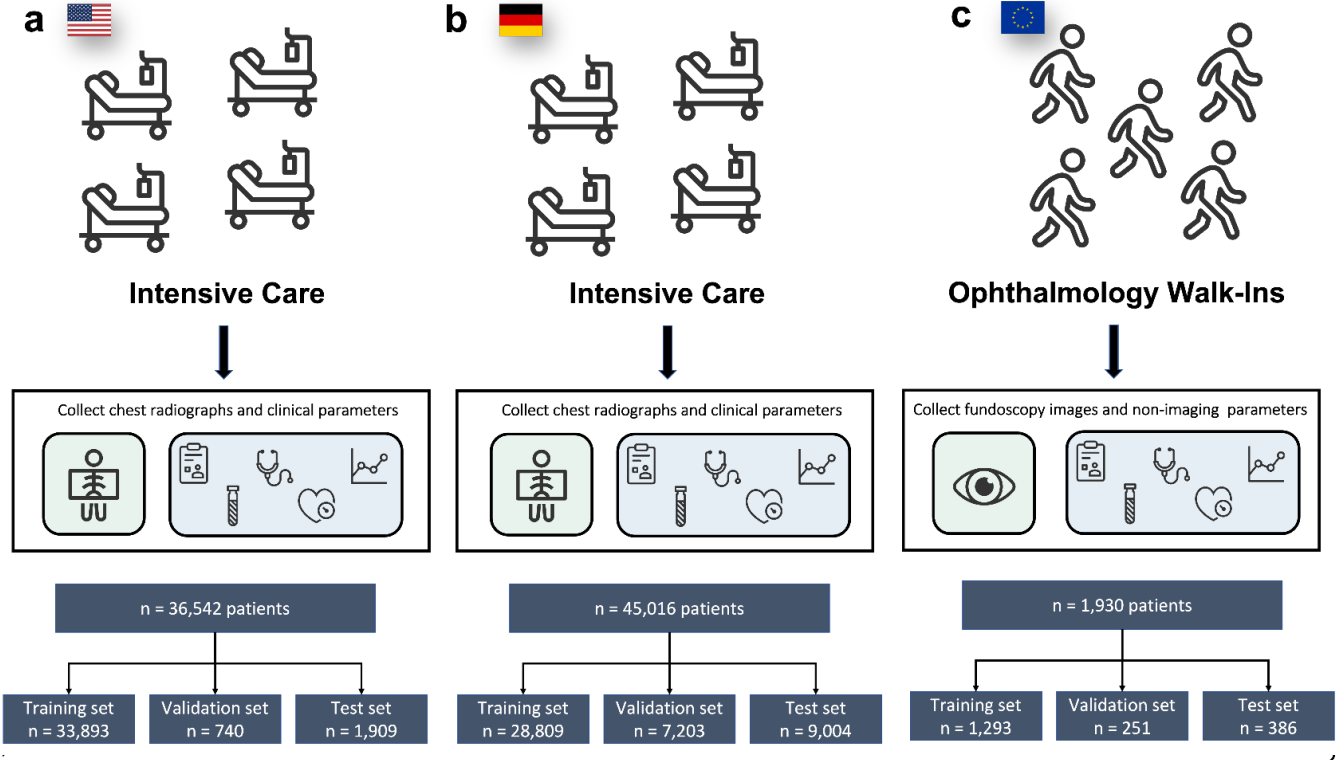

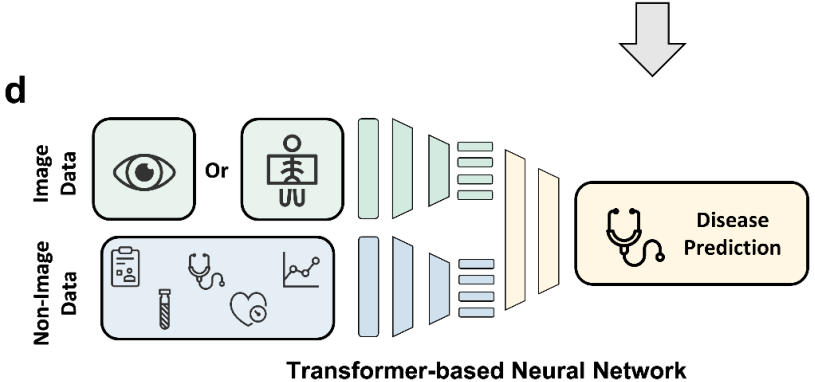

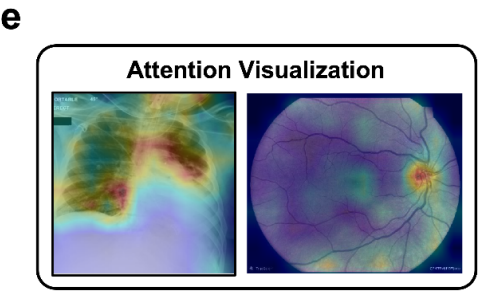

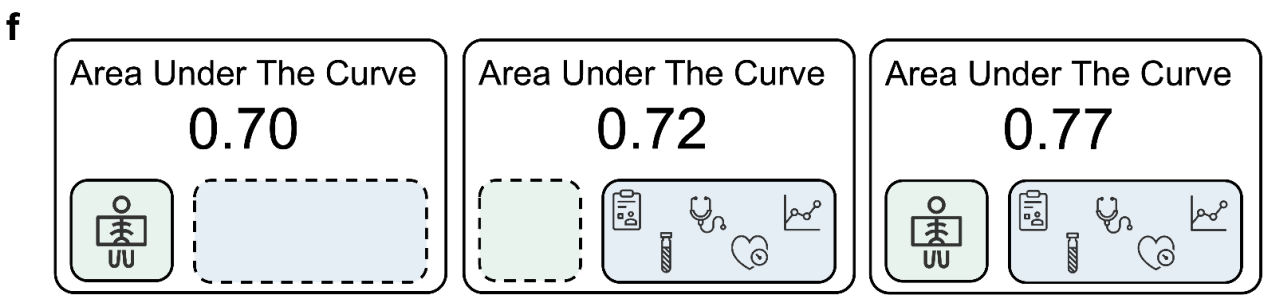

**Figure 1**: Study overview. Imaging and non-imaging information are extracted from three datasets (**a**) the publicly available MIMIC dataset, (**b)** an internal dataset of chest radiographs and accompanying clinical parameters and (**c)** an internal dataset of fundoscopy images from ophthalmology walk-ins. The datasets are then split into training, validation and test sets and (**d)** a transformer-based neural network architecture is trained to predict the dataset-specific outcomes**.** We proceed by leveraging the attention mechanism in the transformer architecture to provide an insight into the decision-making process of the neural network (**e)** and show that the predictive performance of the neural network increases for all three datasets when provided with both, imaging and non-imaging inputs, compared to the unimodal cases **(f)**.

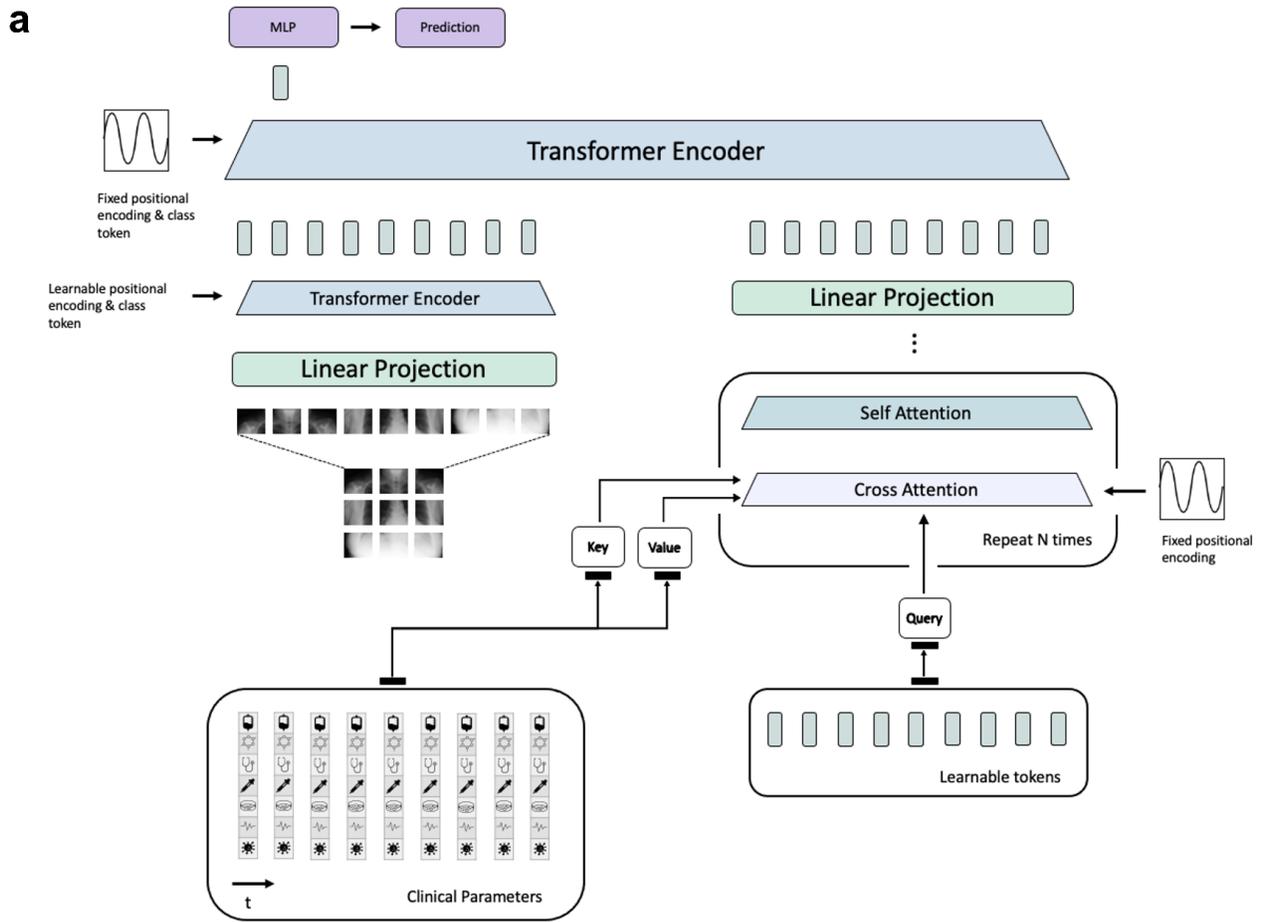

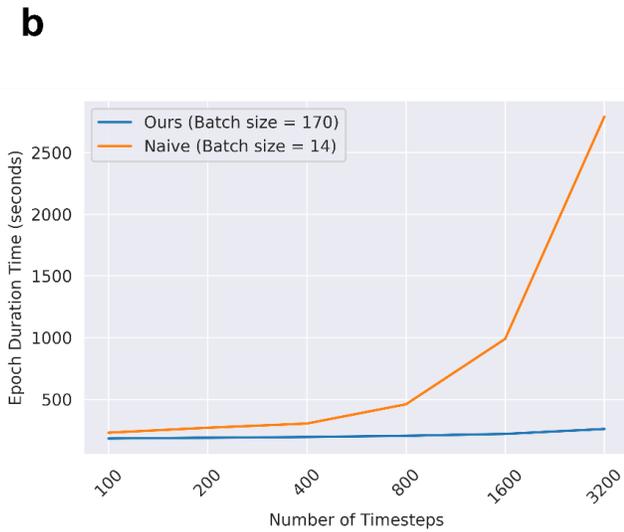
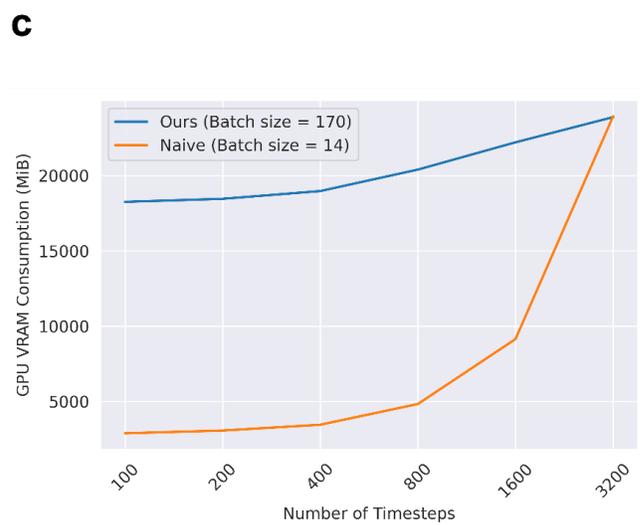

**Figure 2**: **(a)** Visualization of the model architecture. Images are first split into non-overlapping patches and subsequently fed through a transformer encoder. To account for scalability with regards to the number of non-imaging parameters, a fixed set of n=64 learnable tokens serves as the neural networks working memory and cross-attention is employed to feed the clinical information to this working memory.

This keeps the network scalable with respect to the number of input tokens (i.e., clinical parameters). The output tokens of both modality-specific neural networks are then merged in a final transformer encoder, such that information from both modalities is fused. **(b)** Compared to the conventional setting, in which the imaging and non-imaging (time series) data are fed directly into a common transformer encoder block for information fusion, we find that our model results in smaller training times. For a fair comparison we show the epoch duration time when trained on the same GPU (NVIDIA Quadro RTX 6000) and similar GPU VRAM consumption (allows for a batch size of 170 for our model and a batch size of 14 for the naive approach). **(c)** GPU VRAM consumption as a function of the number of input parameters. We find that our approach scales much more efficiently compared to the naive approach for an increasing number of input parameters and therefore allows for larger batch sizes during training. Here, the batch size used for each model was based on the maximal possible batch size (in terms of VRAM consumption of the GPU) when training the model with 3,200 timesteps.

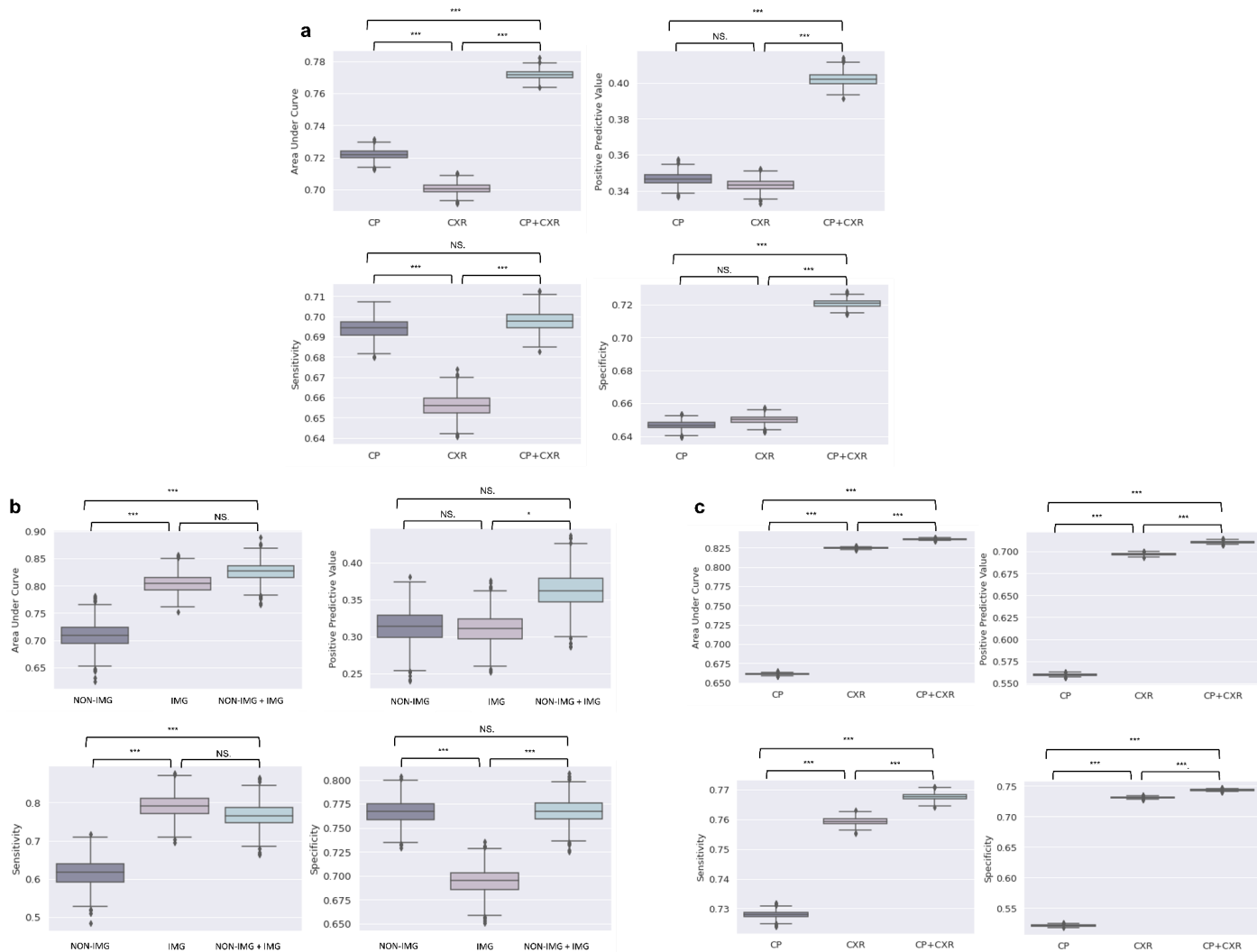

**Figure 3:** Detailed overview of the predictive performance of the trained neural networks in terms of the area under curve (AUC) under the receiver-operating characteristic, positive predictive value, sensitivity and specificity. For each dataset **(a)** MIMIC, **(b)** internal dataset of fundoscopy images and **(c)** internal dataset of chest radiographs, three models were trained that either only relied on the non-imaging information, only on the imaging information or used both modalities as input. The boxplots show the result of n=1,000 bootstrapping runs. In terms of AUC, we find that the multimodal model exhibits a higher score than their unimodal counterparts. Note: NS. denotes a non-significant difference, * denotes P<.05, ** denotes P<.01 and *** denotes P<.001.

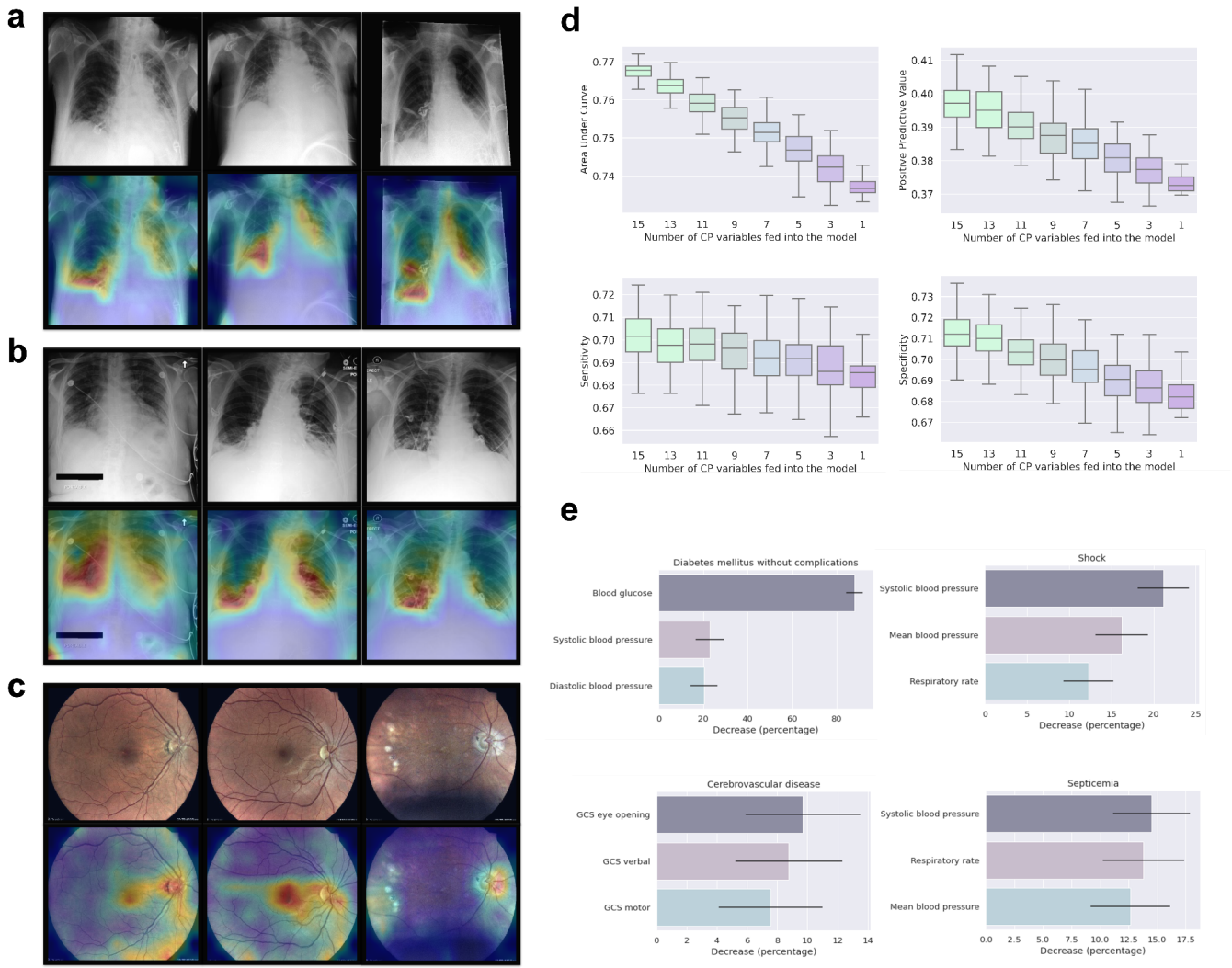

**Figure 4:** Visualization of the explainability studies performed with our neural network architecture. Attention projection of the transformer on the chest radiographs from the local hospital **(a)**, from MIMIC **(b)** and funduscopic images for the screening dataset **(c)**. We find that for all datasets, the transformer model learns to focus on parts of the image that are relevant to human readers, e.g., the optic disc for fundoscopy images and pulmonary opacities for the radiographs. **(d)** Visualizes the performance in terms of area under the curve (AUC) of the receiver operating characteristic, positive predictive value, sensitivity and specificity of the neural network trained on the MIMIC dataset when a number of clinical parameters (non-imaging information) is replaced with null-tokens. Performance continuously drops with an increasing number of dropped clinical parameters. **(e)** To gain understanding of the clinical parameters that mostly affect the performance of the neural network for a specific disease, we visualize the decrease (in percent) in the mutual information between the predicted distribution over all samples

and their ground truth labels when a specific clinical parameter is left out during inference. Error bars denote standard deviation.

# Tables

**Table 1:** Detailed overview over the predictive performance of our model when trained on the publicly available MIMIC dataset. Three different neural networks were trained to predict the presence of 25 different pathologies that either relied entirely on the non-imaging information (CP), the imaging information (CXR) or used both modalities as input (CP+CXR).

|  | AUC Image only (CXR) | AUC Clinical Parameters only (CP) | AUC Imaging and Clinical Parameters (CP+CXR) |
|---|---|---|---|
| **Renal failure** | 0.705 | 0.744 | 0.774 |
| **Cerebrovascular disease** | 0.705 | 0.890 | 0.910 |
| **Myocardial infarction** | 0.684 | 0.713 | 0.764 |
| **Cardiac dysrhythmias** | 0.686 | 0.638 | 0.715 |
| **Kidney disease (chronic)** | 0.714 | 0.704 | 0.783 |
| **Pulmonary disease** | 0.751 | 0.666 | 0.755 |
| **Surgical complications** | 0.649 | 0.697 | 0.721 |
| **Conduction disorders** | 0.835 | 0.665 | 0.852 |
| **Congestive heart failure** | 0.795 | 0.708 | 0.824 |
| **Coronary atherosclerosis** | 0.759 | 0.731 | 0.788 |
| **Diabetes mellitus with complications** | 0.641 | 0.865 | 0.866 |
| **Diabetes mellitus without complication** | 0.634 | 0.731 | 0.734 |
| **Lipid disorders** | 0.680 | 0.667 | 0.716 |
| **Essential hypertension** | 0.637 | 0.645 | 0.693 |
| **Electrolyte disorders** | 0.657 | 0.713 | 0.729 |
| **Gastrointestinal hemorrhage** | 0.634 | 0.739 | 0.755 |
| **Hypertension with complications** | 0.712 | 0.679 | 0.770 |
| **Liver diseases** | 0.698 | 0.679 | 0.739 |
| **Respiratory disease (lower tract)** | 0.610 | 0.585 | 0.613 |
| **Respiratory disease (upper tract)** | 0.596 | 0.724 | 0.726 |
| **Pneumothorax** | 0.760 | 0.628 | 0.757 |
| **Pneumonia** | 0.747 | 0.774 | 0.811 |
| **Respiratory failure** | 0.759 | 0.818 | 0.857 |
| **Septicemia** | 0.727 | 0.794 | 0.805 |
| **Shock** | 0.739 | 0.849 | 0.852 |
| **Mean over all pathologies** | 0.700 | 0.722 | 0.772 |

**Table 2**: Comparison of our method to other state-of-the-art architectures evaluated on the publicly available MIMIC dataset. For comparison we use the area under curve (AUC) of the receiver operating characteristic and compute the average over all available pathologies and 95% confidence intervals.

| Method | Mean AUC | Comments |
| --- | --- | --- |
| Early[28] | 0.739 [0.712, 0.766] | Two feature extractors are utilized for training the non-imaging and imaging data separately. The features are then extracted and concatenated, such that a classification layer can be used to merge the inputs. |
| Joint[28] | 0.754 [0.727, 0.780] | Imaging and non-imaging data are trained end-to-end in a common neural network architecture that consists of two separate feature extraction blocks that are later fused by a final classification head. |
| MMTM[28,33] | 0.734 [0.707, 0.761] | To merge the features of the imaging and non-imaging feature extraction backbone, a Multimodal Transfer Module (MMTM) is used. |
| DAFT[28,34] | 0.737 [0.710, 0.764] | Features from the imaging and non-imaging feature extraction backbone are merged by a Dynamic Affine Feature map Transform (DAFT) block that scales and shifts the feature maps for information fusion. |
| Unified[28,35] | 0.765 [0.742, 0.794] | To merge the features from paired imaging and non-imaging inputs, two feature extractors are employed that in every training iteration are first trained separately on unpaired samples and then in a second step used to extract the features of the paired inputs. The features are concatenated and merged with a final classification head. |
| MedFuse[28] | 0.756 [0.729, 0.782] | Similarly to previous models, two separate feature extractors are trained for the imaging and non-imaging inputs. However, unlike the used classification heads before, LSTM[36] (Long Short-Term Memory) layers are utilized to merge the |
| | | features of both extractors in a sequential manner. |
| MedFuse[28] (OPTIMAL) | 0.770 [0.745, 0.795] | This model is identical to the previous MedFuse model. However, instead of using all unpaired samples, the authors selectively only used a fraction of the unpaired dataset that led to the best performance. |
| Ours (CP+CXR) | **0.772 [0.77, 0.78]** | This model utilizes the transformer architecture to extract features from imaging and non-imaging information. To keep the model scalable to a large number of non-imaging input tokens, the cross-attention module is leveraged and the features from both modalities are fused in a final transformer encoder block. |

# Supplemental Material

**Supplementary Table 1**: Overview of the non-imaging parameters available to the model when trained on the MIMIC dataset. The impute value is used when the clinical parameter is missing and is based on the work from Hayat et al.[28].

| Variable | Type | Missing (%) | Mean (± std) | Impute Value |
|---|---|---|---|---|
| Diastolic blood pressure | Continuous | 0.08 | 60.22 (9.0) mmHg | 59.0 mmHg |
| Fraction inspired oxygen | Continuous | 50.66 | 0.42 (0.07) FiO2 | 0.21 FiO2 |
| Glasgow Coma Scale eye opening | Categorical | 0.42 | 3.64 (0.61) | 4 |
| Glasgow Coma Scale motor response | Categorical | 0.45 | 5.29 (1.43) | 6 |
| Glasgow Coma Scale verbal response | Categorical | 0.44 | 4.44 (1.11) | 5 |
| Glucose | Continuous | 2.30 | 128.59 (27.04) mg/dL | 128.0 mg/dL |
| Heart rate | Continuous | 0.01 | 83.41 (12.07) bpm | 86 bpm |
| Body height | Continuous | 98.46 | 170.60 (8.63) cm | 170.0 cm |
| Mean blood pressure | Continuous | 0.09 | 76.12 (9.90) mmHg | 77.0 mmHg |
| Oxygen saturation | Continuous | 0.10 | 97.53 (1.96) % | 98.0 % |

| Respiratory rate | Continuous | 0.07 | 18.93 (3.76) breaths per minute | 19 breaths per minute |
| --- | --- | --- | --- | --- |
| Systolic blood pressure | Continuous | 0.08 | 116.59 (15.14) mmHg | 118.0 mmHg |
| Temperature | Continuous | 0.81 | 36.87 (0.32) °C | 36.6 °C |
| Body weight | Continuous | 12.21 | 81.70 (15.11) kg | 81.0 kg |
| pH | Continuous | 32.69 | 7.39 (0.06) | 7.4 |

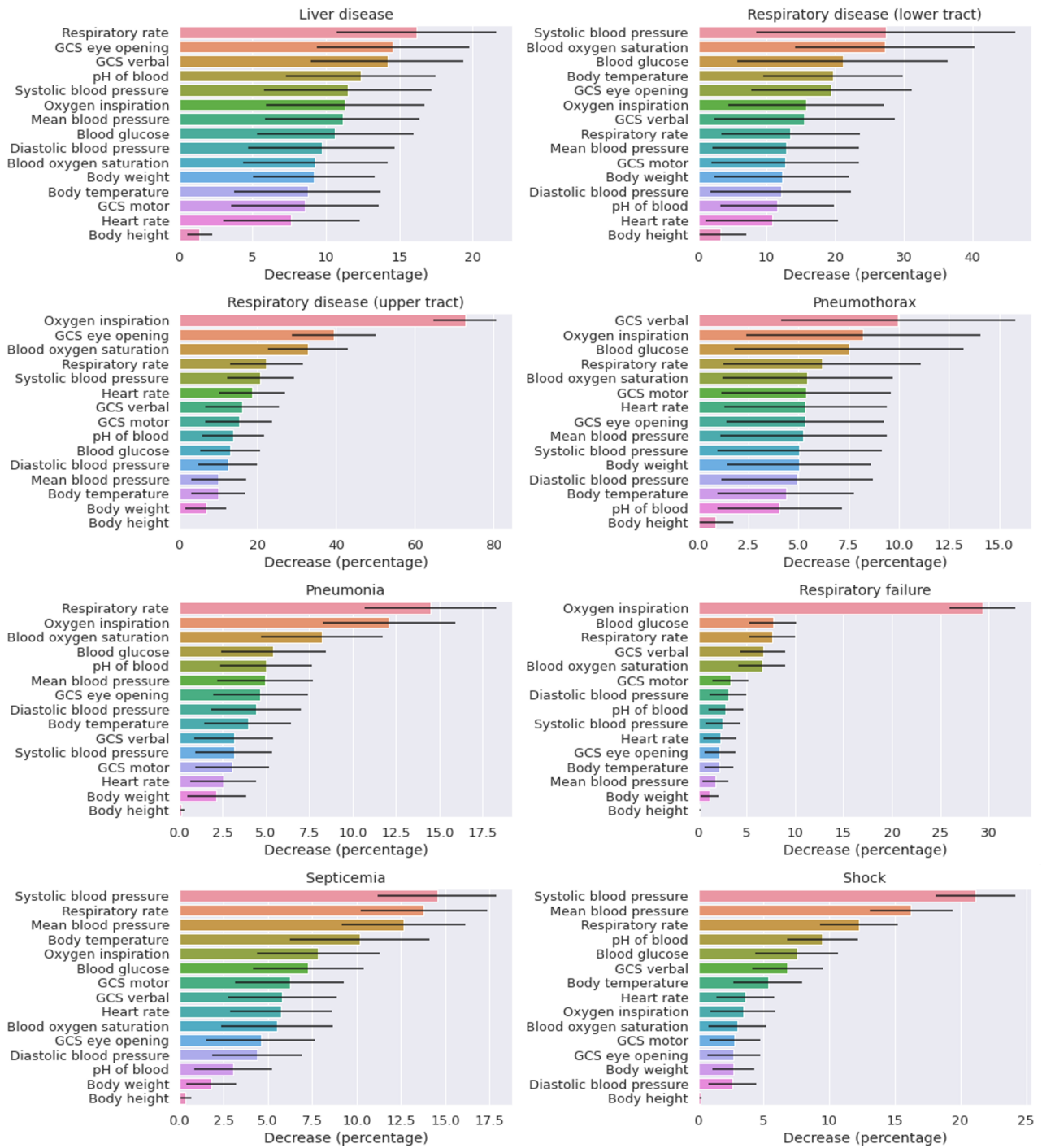

**Supplementary Figure 1:** Overview of the loss in mutual information when leaving out a specific non-imaging parameter for all pathologies available in the used MIMIC dataset.

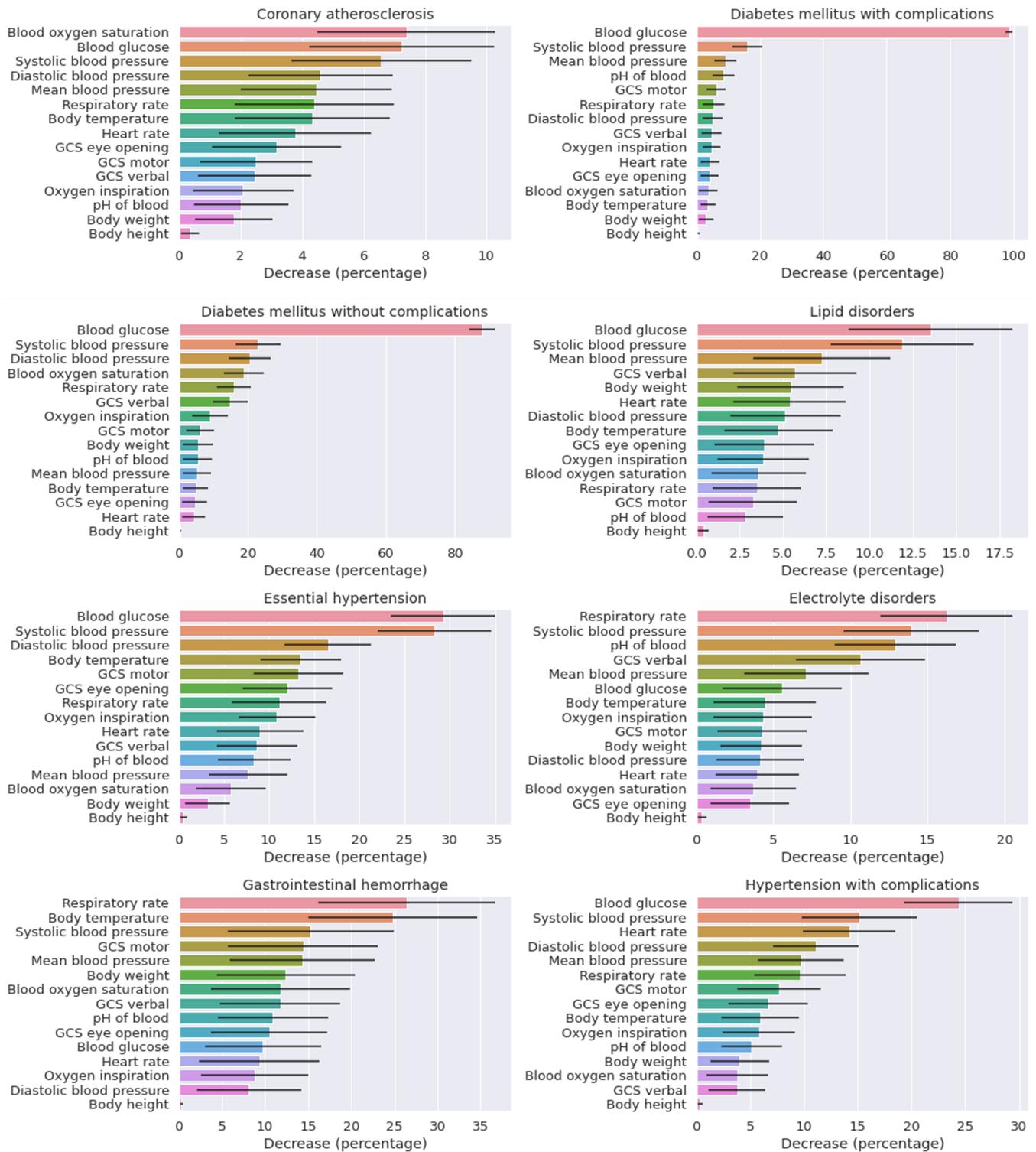

**Supplementary Figure 1:** Continued.

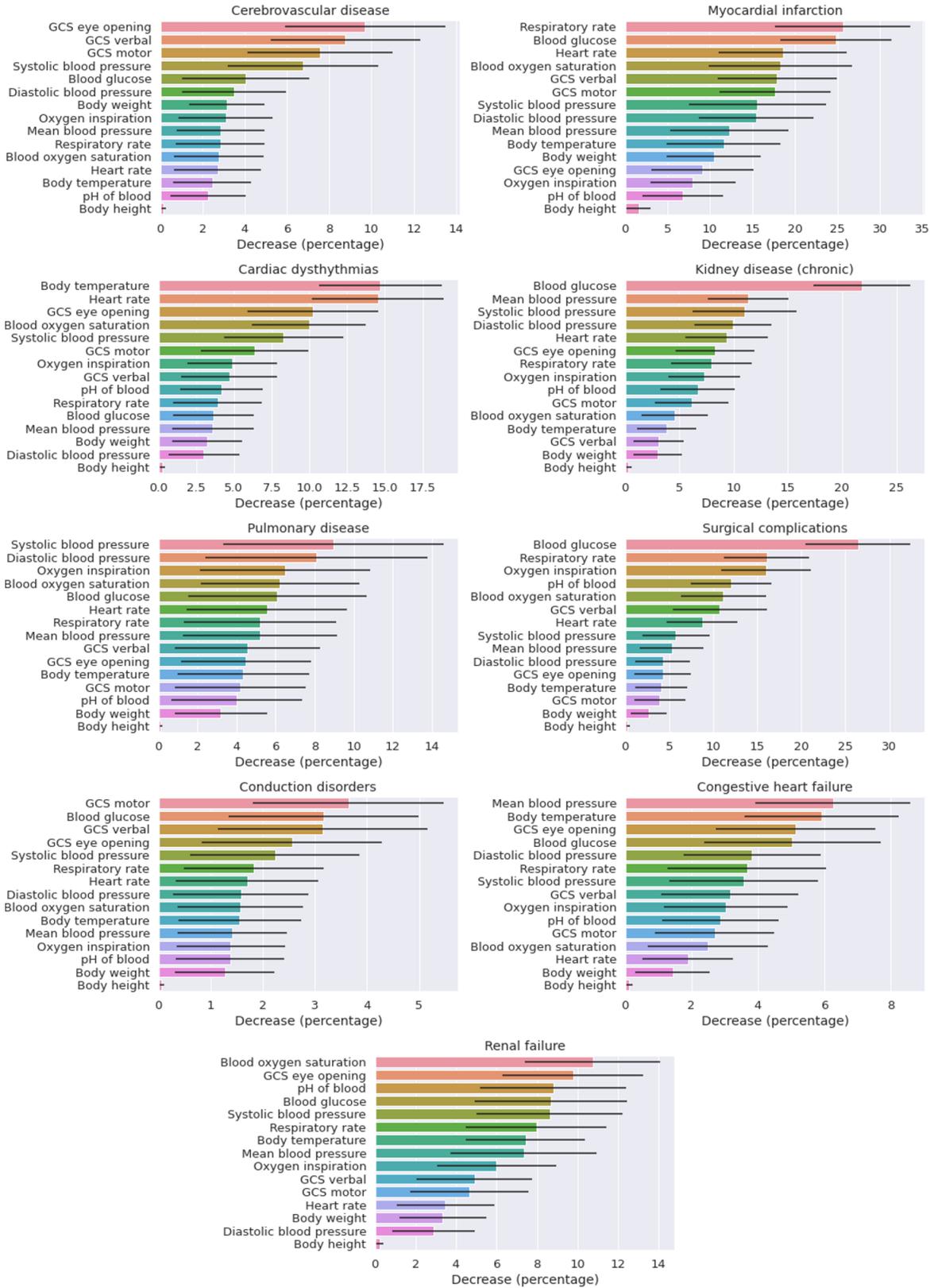

**Supplementary Figure 1:** Continued.